\pdfoutput=1

\documentclass[11pt]{article}

\usepackage{acl}

\usepackage{amssymb}
\usepackage{times}
\usepackage{latexsym}
\usepackage{graphicx}
\usepackage{amsmath}
\usepackage{amsfonts}
\usepackage{multirow}
\usepackage{booktabs}
\usepackage{stmaryrd}
\usepackage{subfigure}
\usepackage{subcaption}
\usepackage{pgfplots}
\usepackage{makecell}
\usepackage{xcolor}
\usepackage{multicol}
\usepackage{booktabs}
\usepackage{bashful}

\definecolor{mycolor_green}{HTML}{D5E8D4}
\definecolor{mycolor_red}{HTML}{F8CECC}

\usepackage[T1]{fontenc}

\usepackage[utf8]{inputenc}

\usepackage{microtype}

\usepackage{inconsolata}

\newcommand*\samethanks[1][\value{footnote}]{\footnotemark[#1]}

\title{\emph{Both Matter}: Enhancing the Emotional Intelligence of Large Language Models without Compromising the General Intelligence}

\author{Weixiang Zhao$^1$, Zhuojun Li$^1$\thanks{\ \ \ Equal contribution}\, Shilong Wang$^1$\samethanks, Yang Wang$^1$, Yulin Hu$^1$, \\ \textbf{Yanyan Zhao}$^1$\thanks{\ \ Corresponding author}, \textbf{Chen Wei}$^2$, \textbf{Bing Qin}$^1$ \\
        $^1$Harbin Institute of Technology, Harbin, China\\
        $^2$Huawei Inc., Beijing, China\\
        \texttt{\{wxzhao, yyzhao, qinb\}@ir.hit.edu.cn}}

\begin{document}
\maketitle
\begin{abstract}
Emotional Intelligence (EI), consisting of emotion perception, emotion cognition and emotion expression, plays the critical roles in improving user interaction experience for the current large language model (LLM) based conversational general AI assistants. Previous works mainly focus on raising the emotion perception ability of them via naive fine-tuning on EI-related classification or regression tasks. However, this leads to the incomplete enhancement of EI and catastrophic forgetting of the general intelligence (GI). To this end, we first introduce \textsc{EiBench}, a large-scale collection of EI-related tasks in the text-to-text format with task instructions that covers all three aspects of EI, which lays a solid foundation for the comprehensive EI enhancement of LLMs. Then a novel \underline{\textbf{Mo}}dular \underline{\textbf{E}}motional \underline{\textbf{I}}ntelligence enhancement method (\textbf{MoEI}), consisting of Modular Parameter Expansion and intra-inter modulation, is proposed to comprehensively enhance the EI of LLMs without compromise their GI. Extensive experiments on two representative LLM-based assistants, Flan-T5 and LLaMA-2-Chat, demonstrate the effectiveness of MoEI to improving EI while maintain GI. \footnote{Our source code is available at \url{https://github.com/circle-hit/MoEI}.}
\end{abstract}

\section{Introduction}
\begin{flushright}
\begin{quote}
    \emph{The question is not whether intelligent machines can have any emotions, but whether machines can be intelligent without any emotions.}\\ 
    \textit{\hfill -- Marvin Minsky}
\end{quote}
\end{flushright}

Emotional intelligence (EI), a pivotal concept in the field of human intelligence, holds significant importance in the context of the current large language models (LLMs) \citep{brown2020language,raffel2020exploring,touvron2023llama} exhibiting great general intelligence (GI) to serve as conversational general AI assistants \citep{minsky2007emotion}. It involves effectively deal with EI-related downstream tasks to accurately perceive, understand users' emotional states and respond properly, which is necessary for fostering effective communication and facilitating smooth social interactions \citep{mayer2001emotional}.

\begin{figure}
\centering
\includegraphics[width=1\columnwidth]{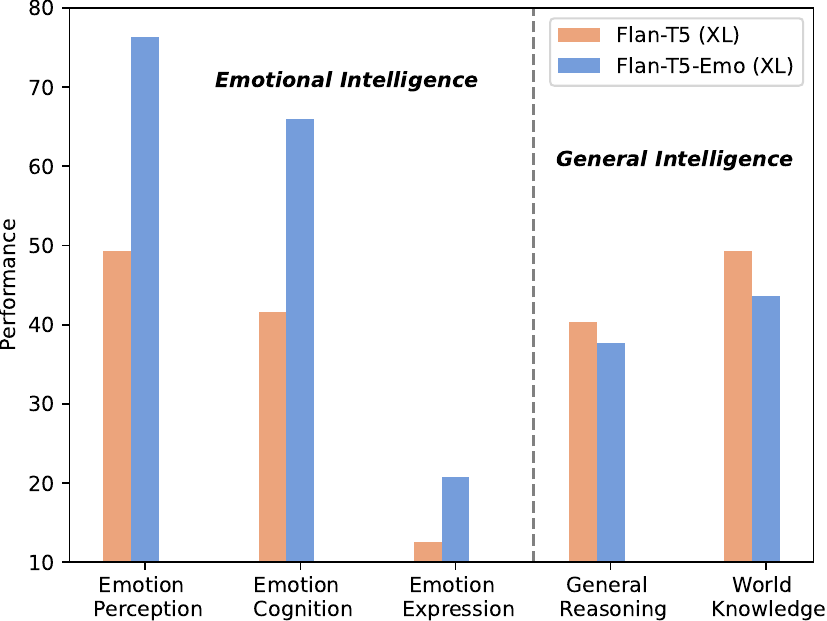}
\caption{Comparison of EI and GI before (in orange) and after (in blue) the EI-enhancement via naive fine-tuning based on the Flan-T5-XL (3B) backbone.}
\label{example}
\end{figure}

To enhance the EI, many researchers have started performing the naive fine-tuning of LLMs on EI-related tasks \citep{zhang2023enhancing,zhang2023dialoguellm,lei2023instructerc,liu2024emollms}. However, there are two significant limitations in the current state of these efforts.

\textbf{On one hand}, the EI-enhancement in these studies is of narrow scope because they only focus on emotion perception regarding emotion classification tasks. However, EI is a broad concept that also includes emotion cognition (e.g. emotion cause reasoning \citep{poria2021recognizing}) and expression (e.g. empathethic response generation \citep{rashkin2018towards}). Therefore, the comprehensive enhancement of EI for LLMs is ignored in previous works.


\textbf{On the other hand}, they all overlook the catastrophic forgetting \citep{mccloskey1989catastrophic} of the GI of LLM backbones during the process of enhancing EI via naive fine-tuning. In our preliminary experiments depicted in Figure \ref{example}, we find that solely focusing on improving the EI of an LLM backbone would result in a significant decline in its GI, such as world knowledge and general reasoning. Desirable intelligent assistants are supposed to be of both high GI and high EI, where any performance loss on either front would significantly undermine the user experience.

To this end, we study \emph{how to comprehensively enhance the EI of current LLM-based assistants while preserving their inherent GI from being compromised}. According to three aspects of EI defined by \citet{mayer2001emotional}, named emotion perception, emotion cognition and emotion expression, we first construct \textsc{EiBench}, a large curated collection of EI-related tasks converted into a text-to-text format, covering 15 tasks with 88 datasets. Moreover, motivated by the promising gains from fine-tuning with task instructions \cite{sanh2021multitask,wang2022super,longpre2023flan}, we also manually write related instructions for each dataset. Thus, \textsc{EiBench} lays a solid foundation for the comprehensive EI enhancement of LLMs.


Further, we propose a novel \underline{\textbf{Mo}}dular \underline{\textbf{E}}motional \underline{\textbf{I}}ntelligence enhancing method (\textbf{MoEI}) with two collaborative techniques, Modular Parameter Expansion (MPE) and Intra-Inter Modulation (I\textsuperscript{2}M), to comprehensively improve the EI of LLMs while maintaining most of their GI. Specifically, in MPE, a set of modular parameters are introduced to endow additional capacity to handle various tasks within the above three aspects of EI. For the sake of computation and resource efficiency, these expanded modular parameters are instantiated with parameter-efficient LoRA blocks \citep{hu2021lora}, named MoLoRA. During the process of EI-enhancement, only the EI-specific MoLoRA is updated, thereby reducing the impact on parameters of the LLM backbone representing its GI. In addition, a router is devised in I\textsuperscript{2}M to exert the modulation on the two separate parameters. More specifically, intra-modulation is performed within MoLoRA, leveraging the weighted combination of different LoRA blocks to deal with various EI-related tasks. And inter-modulation functions to strike the balance between MoLoRA and the whole LLM backbone to achieve the goal of protecting GI, where the GI-related samples are navigated to be processed by the LLM backbone while reducing the influence from EI-specific MoLoRA.

We conduct extensive experiments on two representative open-source LLM-based assistants, Flan-T5 \citep{chung2022scaling} and LLaMA-2-Chat \citep{touvron2023llama}. Results demonstrate that MoEI not only helps significantly enhance all three aspects of their EI but also ensures that their GI, including world knowledge, general reasoning, commonsense reasoning and reading comprehension, are hardly compromised. Moreover, such EI-enhanced models could exhibit better performance when addressing various OOD tasks in the presence of emotional stimuli \citep{li2023large}.

The main contributions of this work are summarized as follows: (1) We take the first step to study how to develop an LLM-based assistant that possesses both high GI and EI, a challenging direction for the more practical application of LLMs. (2) We introduce \textsc{EiBench}, a comprehensive collection of EI-related tasks to support the EI enhancement of LLM backbones. Then a novel method MoEI is proposed to comprehensively improve the EI of LLM backbones without compromising their GI. (3) Experiments on various EI and GI benchmarks demonstrate the effectiveness of MoEI.

\begin{figure}
\centering
\includegraphics[width=0.8\columnwidth]{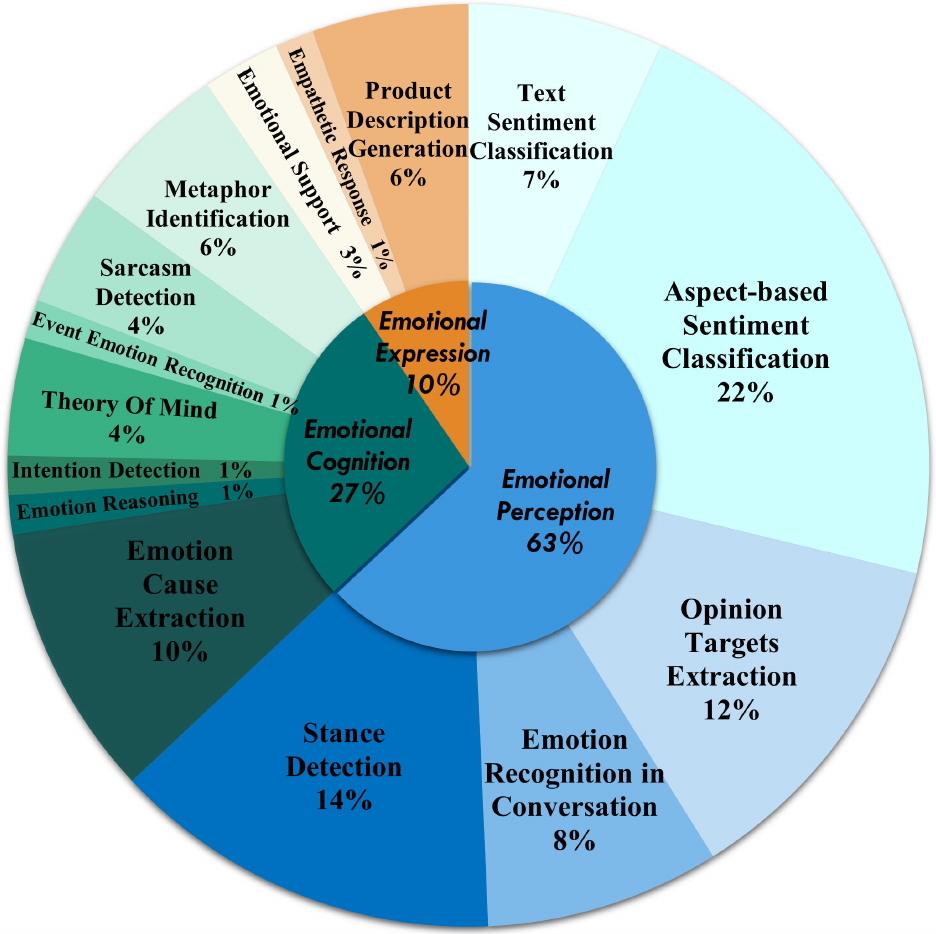}
\caption{Overview of \textsc{EiBench}. 15 EI-related tasks are categorized into 3 main categories: emotion perception, emotion cognition, and emotion expression.}
\label{eibench}
\end{figure}

\section{\textsc{EiBench}}
We first introduce \textsc{EiBench}, a large-scale collection of EI-related tasks with instructions that describe them in plain language. Figure \ref{eibench} shows the overview of the benchmark by category and task.

\paragraph{Taxonomy of Emotional Intelligence.} According to \citet{mayer2001emotional}, the EI of LLMs includes three aspects: (1) Emotion Perception, (2) Emotion Cognition and (3) Emotion Expression. Please refer to Appendix \ref{taxonomy_ei} for their specific meanings.

\paragraph{Task Collection.} Based on the aforementioned EI taxonomy, we systematically gather and organize existing open-source tasks and datasets related to EI within this framework. The resulting depiction of all EI-related tasks in \textsc{EiBench} is illustrated in Figure \ref{eibench}, while Table \ref{all_task} (in Appendix \ref{task_ei}) provides a comprehensive list of datasets associated with each task. This leads to 15 tasks and 88 datasets in total. To be more specific, in the aspect of emotion perception, the primary focus is on classification tasks such as emotion recognition and stance detection. Shifting to the area of emotion cognition, the prioritized task involves extracting emotion causes. Additionally, we incorporate more advanced cognitive challenges, including irony and metaphor recognition, aiming to comprehensively enhance the LLMs' emotional cognitive capabilities. Finally, in the realm of emotion expression, the main tasks involve empathetic response and emotional support, empowering the model to offer improved comfort and guidance to users, thereby enhancing the overall interactive experience.

\paragraph{Task Schema.} Motivated by the promising gains from fine-tuning with task instructions \cite{sanh2021multitask,wang2022super,longpre2023flan}, we manually construct one piece of instruction for each dataset in our \textsc{EiBench}. Specifically, the schema contains the following components: (1) \textbf{Text Input:} the input sentence $X$. (2) \textbf{Instruction:} the detailed guidance on how the model should process $X$ to complete the current task. (3) \textbf{Option} (for classification task only): including all the candidate labels and serving as both a constraint and a hint. We hire 5 annotators who are proficient in English to write these instructions. In addition, to guarantee the quality of these instructions, 1 or 2 reviewers are also assigned to each dataset. Their task is to confirm that whether the instructions are clear, fluent and comprehensive enough for an average language speaker to successfully complete the given task. Examples of instances from our \textsc{EiBench} is displayed in Table \ref{task_instance} in Appendix \ref{task_ei}.

\section{Methodology}
\begin{figure}
\centering
\includegraphics[width=1\columnwidth]{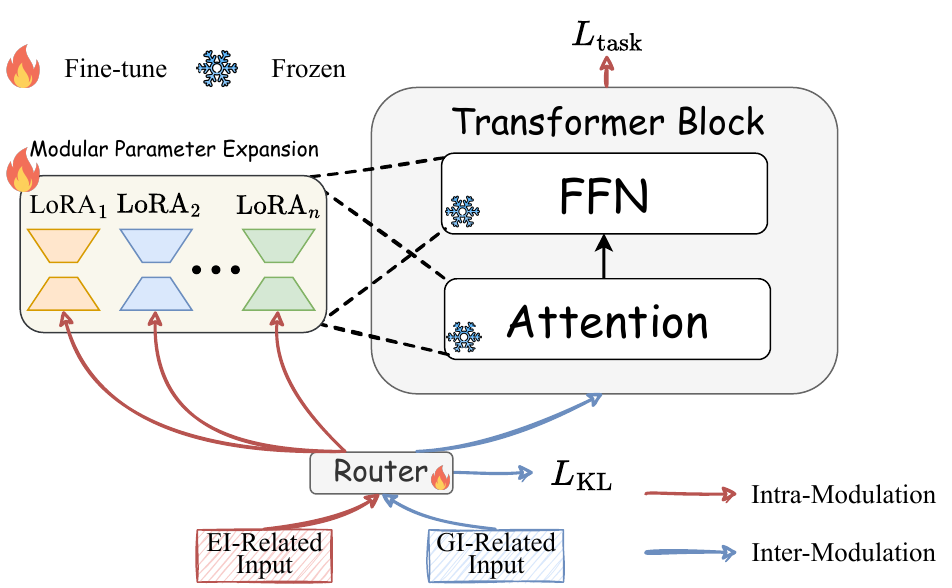}
\caption{The overall architecture of our proposed MoEI framework, which consists of two techniques, modular parameter expansion and intra-inter modulation. Red and blue lines represent the forward flow of the EI- and GI related inputs that participate in the Intra- and Inter-Modulation, respectively.}
\label{moei}
\end{figure}

Instead of merely focusing on enhancing specific capabilities of LLMs, we present a novel approach MoEI, a model-agnostic EI-enhancing method that is compatible with any transformer-based LLM, which could not only comprehensively boost the EI of LLMs but also safeguards their GI from being compromised. As shown in Figure \ref{moei}, MoEI consists of two collaborative techniques, namely Modular Parameter Expansion and Intra-Inter Modulation. The subsequent section will offer a detailed introduction to both of them.

\subsection{Preliminary}
\paragraph{Low Rank Adaptation.} We adopt a representative PET method LoRA \citep{hu2021lora} in MoEI. Specifically, in LoRA, the pre-trained weight matrix of LLMs is expanded with a low-rank decomposition. For any linear layer $h=W_0x$, the forward pass with LoRA is modified to be:
\begin{equation}
    h=W_0x+BAx
\end{equation}
where $W_0 \in \mathbb{R}^{d\times k}$, $B \in \mathbb{R}^{d\times r}$, $A \in \mathbb{R}^{r\times k}$ and the rank $r \ll \min(d,k)$. The pre-trained weight $W_0$ remains fixed during training, while $A$ and $B$ are trainable parameters.

\subsection{Modular Emotional Intelligence Enhancement}

\paragraph{Modular Parameter Expansion.} To harness two advantages of LoRA, which includes isolation from the entire parameters of LLMs and promising training efficiency, we take a step further by extending it to modular designs, endowing more capacity to accomodate various newly acquired EI-related knowledge. This leads to the Mixture-of-LoRA (MoLoRA) architecture.

MoLoRA inherits the flexibility of LoRA and could be applied on any linear layer of the transformer-based LLM backbone (mainly in the attention and FFN layer) to expand the model capacity with multiple pairs of low-rank matrixes. To be more specific, the singular pair of low-rank matrixes $BA$ in Equation (1) is now replaced by a set containing multiple ones $\{B_iA_i\}^N_{i=1}$ and the forward process of the MoLoRA layer can be mathematically expressed as follows:
\begin{equation}
    h=W_0x + \sum_{i=1}^N B_iA_ix
\end{equation}
where $N$ is the number of modular LoRA blocks.

\paragraph{Intra-Inter Modulation.} To effectively navigate different types of inputs to be properly processed by its corresponding parameters, a router is introduced in I\textsuperscript{2}M to exert the intra- and inter-modulation on weighted contributions of the separate MoLoRA and the LLM backbone:
\begin{equation}
\begin{aligned}
    h = \alpha &W_0x + \sum_{i=1}^N \beta_i \ B_iA_ix \\
     [\alpha; \{\beta_i\}_{i=1}^N] &=  G(x) = \textrm{softmax} (W x)
\end{aligned}
\end{equation}
where $G(\cdot)$ represents the router and $W \in \mathbb{R}^{d \times {N+1}}$ is the trainable parameter of it. $\alpha$ and $\{\beta_i\}_{i=1}^N$ are the intensity weights of inter and intra-modulation, respectively. In the upcoming contents, we will elaborate the concept and delve into the optimization process associated with them.

More specifically, intra-modulation $\{\beta_i\}_{i=1}^N$ is only performed within EI-specific MoLoRA to leverage the weighted combination of different LoRA blocks, which leads to more powerful capability in dealing with various EI-related tasks compared to the single LoRA. As shown in Figure \ref{moei}, the optimization of intra-modulation is driven by the EI-related inputs with the task loss:
\begin{equation}
L_{\text{task}} =-\sum_{(x, y) \in \mathcal{EI}} \log P\left(y \mid x; \theta_m, \theta_E, \theta_G \right)
\end{equation}
where $\theta_m, \theta_E$ and $\theta_G$ are parameters of the LLM backbone, the MoLoRA and the router, respectively. $\mathcal{EI}$ is the training data samples from \textsc{EiBench}. Only parameters of $\theta_E$ and $\theta_G$ are updated during the training.

And inter-modulation $\alpha$ functions to strike the balance between MoLoRA and the whole LLM backbone, ensuring the GI-related samples to be processed only by the LLM backbone and eliminating the influence from EI-specific MoLoRA. This is achieved through the minimizing of a KL divergence loss:
\begin{align} \label{kl_loss}
\small
L_{\text{KL}} = \sum_{(x, y) \in \mathcal{GI}} D_{\text{KL}} (G(x) || I)
\end{align}
where $\mathcal{GI}$ is the replayed GI-related samples from the previous training corpus of the LLM backbone, which is a subset of Flan collection \citep{longpre2023flan} in our experiments. $I$ is the one-hot vector with only the position of $\alpha$ setting to 1.

And it is worth to mention that, in our experiments, although $\alpha$ does involve in the softmax function, it is exactly set to 1 in the forward pass because we believe the large-scale knowledge stored in the contemporary LLMs are the key to handle downstream tasks, which is also verified by our main experiments in Table \ref{main results}. In general, the inter-modulation ensures the given input, either EI- or GI-related, to fully leverage the powerful LLM backbone, while intra-modulation determines to what extent the incremental EI-enhanced MoLoRA are activated to complete the current input.


Finally, a multi-task learning fashion is adopted to jointly minimize the task loss and the KL loss:
\begin{equation}
L= L_{\text{task}}+\lambda L_{\text{KL}}
\end{equation}
where $\lambda$ functions to balance the two parts.

\begin{table*}[h!]
\centering
\resizebox{\linewidth}{!}{
\begin{tabular}{l |c c c |c c c c }
\toprule

\textbf{}        & \multicolumn{3}{c|}{\textbf{Emotional Intelligence}} & \multicolumn{4}{c}{\textbf{General Intelligence}} \\
\textbf{}        & \textbf{Emo.Prc}      & \textbf{Emo.Cog} & \textbf{Emo.Exp}      & \textbf{WK}        & \textbf{GR}      & \textbf{CR}       & \textbf{RC}     \\ \midrule
Flan-T5     & 49.35    & 41.66   & 12.60 & 49.36 &40.38 &68.88 &87.49    \\
Flan-T5 FT     &76.27 &\colorbox{mycolor_red}{65.94} &20.81 & \colorbox{mycolor_red}{43.64} & 37.67 & 68.39 & 86.91    \\
+ Replay  &  76.38      &  \colorbox{mycolor_green}{69.36}   &  \colorbox{mycolor_red}{18.79}   & 44.35  &36.23 &  68.23 & 87.09    \\
Flan-T5 LoRA     & \colorbox{mycolor_red}{76.11} & 68.18 & 21.05 & 47.39 & 34.60 & \colorbox{mycolor_red}{68.21} &  \colorbox{mycolor_red}{86.85}  \\
+ Replay  &  76.31 & 67.41 & 21.09   & 45.76 & \colorbox{mycolor_red}{32.93} & 68.72 & 87.28    \\
\midrule
\textbf{Flan-T5 MoEI} (Ours)  & \colorbox{mycolor_green}{77.15} & 68.32 & \colorbox{mycolor_green}{25.02}  & \colorbox{mycolor_green}{49.23} & \colorbox{mycolor_green}{40.58} & \colorbox{mycolor_green}{68.99} & \colorbox{mycolor_green}{87.61}   \\
\midrule
\midrule
LLaMA-2-Chat     &  19.81  & 24.88  & 11.89 & 46.97    & 33.56   & 76.44 & 79.76  \\
LLaMA-2-Chat FT    & \colorbox{mycolor_red}{46.31}   & 45.01  & 11.58 & \colorbox{mycolor_red}{24.28}   & 14.07  & 58.87 & 59.88   \\
+ Replay  &  47.65  &  \colorbox{mycolor_red}{39.67} & \colorbox{mycolor_red}{11.40} &  24.38  & \colorbox{mycolor_red}{3.66}   &  \colorbox{mycolor_red}{57.24} &  \colorbox{mycolor_red}{59.20}  \\
LLaMA-2-Chat LoRA   & 74.62   & 65.78  & 19.39 & 36.73   & 31.07   & 75.84 &  78.59  \\
+ Replay  &  74.96  & 66.75  & 18.96 &  41.33  & 28.44   & 61.15 & 75.41   \\
\midrule
\textbf{LLaMA-2-Chat MoEI} (Ours)  &  \colorbox{mycolor_green}{76.85}  & \colorbox{mycolor_green}{68.93}   & \colorbox{mycolor_green}{21.01}  &  \colorbox{mycolor_green}{46.15}   &  \colorbox{mycolor_green}{35.56}  & \colorbox{mycolor_green}{78.35} & \colorbox{mycolor_green}{81.13} \\
\bottomrule
\end{tabular}
}
\caption{The overall results on the EI and GI benchmarks with Flan-T5-XL (3B) and LLaMA-2-Chat-7B backbone. The best and worst results are signaled by the green and red background, respectively.}
\label{main results}
\end{table*}

\section{Experiments}
\subsection{Dataset and Evaluation Metrics}
\paragraph{Emotional Intelligence.} We split \textsc{EiBench} into two subsets: one for training and the other for evaluation. For the training set, we sample a maximum of 5,000 instances from each dataset, resulting in 268,234 training instances in $\mathcal{EI}$. For an efficient evaluation, a maximum of 100 instances are sampled from the remaining sets of each dataset, leading to 5,600 instances as the evaluation under the supervised settings. We report the average accuracy for datasets from Emotion Perception (Emo.Prc) and Rouge-L \citep{lin2004rouge} for those from Emotion Cognition (Emo.Cog) and expression (Emo.Exp). Furthermore, to comprehensively assess the impact of EI-enhancement, we extend our evaluation to include EQ-Bench \citep{paech2023eq} for cross-task zero-shot evaluation, where models are tasked with predicting the intensity of emotions of characters in a dialogue with a set of 60 English questions.

\paragraph{General Intelligence.} To evaluate a model’s GI, following prior research \citep{wang2023trace}, we conduct evaluations across four crucial dimensions: (1) World Knowledge (WK): Employing the Massive Multitask Language Understanding dataset (MMLU) \citep{hendrycks2020measuring}, with questions spanning 57 subjects, ranging from elementary to professional levels. Following LLaMa-2 \citep{touvron2023llama}, we report 5-shot accuracy. (2) General Reasoning (GR): The evaluation involves Big-Bench-Hard (BBH) \citep{suzgun2022challenging}, featuring 23 tasks derived from Big-Bench \citep{ghazal2013bigbench}. Few-shot prompting, provided with 3-shot in-context examples, are utilized and EM scores are reported. (3) Commonsense Reasoning (CR): Our assessment incorporates PIQA \citep{bisk2020piqa}. Following LLaMa-2 \citep{touvron2023llama}, we report 0-shot accuracy. (4) Reading Comprehension (RC): BoolQ \citep{clark2019boolq} is adopted for evaluating reading comprehension, with the focus on reporting 0-shot accuracy.

\begin{figure}
\centering
\includegraphics[width=1\columnwidth]{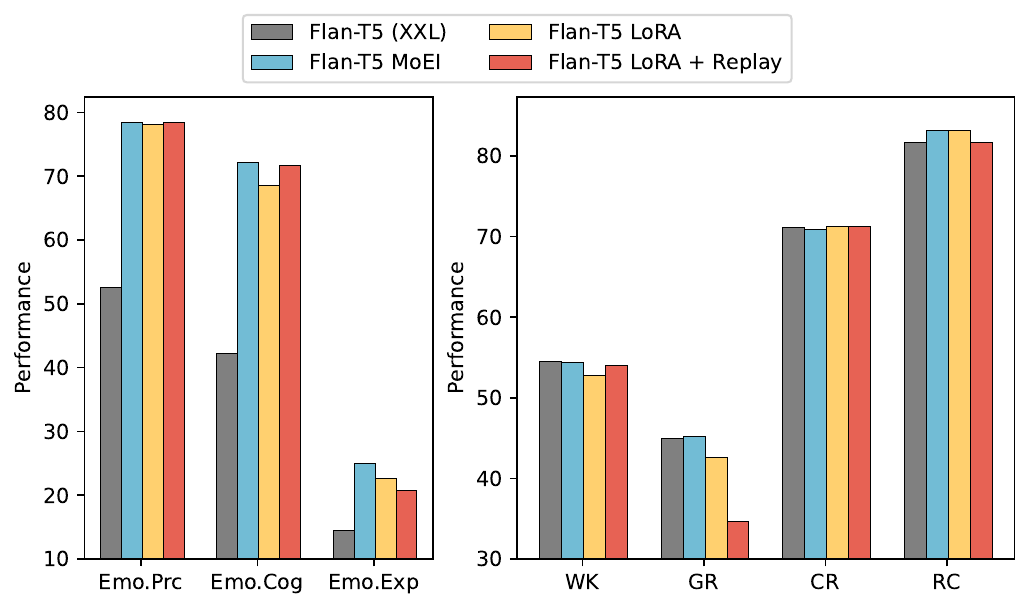}
\caption{Results of EI and GI of different methods on the larger Flan-T5-XXL (11B) backbone.}
\label{scale}
\end{figure}

\subsection{Baselines and Comparison Models}
Based on two representative LLM-based assistants of different architectures, Flan-T5 (encoder-decoder) \citep{chung2022scaling} and LLaMA-2-Chat (decoder-only) \citep{touvron2023llama}, we evaluate MoEI against the following EI-enhancing methods: (1) \textbf{FT}: directly fine-tunes the LLM backbone with the whole parameters of it updated. (2) \textbf{LoRA} \citep{hu2021lora}: updates additional parameter-efficient low rank matrixes and weights of the LLM backbone remain frozen, which could be viewed as the method of singular parameter expansion without intra-inter modulation.
Moreover, both baseline methods are seamlessly integrated with \textbf{Replay} \citep{lopez2017gradient}, a widely adopted continual learning method to mitigate the catastrophic forgetting of previous knowledge, where the model is trained with a mixture of EI and GI data in the multi-task learning fashion.

\subsection{Implementation Details}
In MoEI, the low rank $r$ of MoLoRA is 4 and $N$ is 8. And the $r$ of the single LoRA-tuning is 32 for the fair comparison. MoEI and LoRA are both implemented upon the query and value projection in the attention layer, and the last linear transformation in the FFN layer. All methods are trained for 5 epochs with 3 random runs. For more detailed settings, please refer to the Appendix \ref{implementation}.

\begin{table*}[h!]
\centering
\resizebox{\linewidth}{!}{
\begin{tabular}{l |c c c |c c c c }
\toprule
\textbf{}        & \multicolumn{3}{c|}{\textbf{Emotional Intelligence}} & \multicolumn{4}{c}{\textbf{General Intelligence}} \\
\textbf{}        & \textbf{Emo.Prc}      & \textbf{Emo.Cog} & \textbf{Emo.Exp}      & \textbf{WK}        & \textbf{GR}      & \textbf{CR}       & \textbf{RC}     \\ \midrule
\textbf{Flan-T5 MoEI}  & 77.15 & 68.32 & \colorbox{mycolor_green}{25.02}  & 49.23 & 40.58 & \colorbox{mycolor_green}{68.99} & \colorbox{mycolor_green}{87.61}   \\
\midrule
-- Modular Expansion & \colorbox{mycolor_red}{75.31} & 66.42 & \colorbox{mycolor_red}{20.29} &  49.10  &  40.18  & 68.19 & 87.52   \\
-- Intra-Modulation &  76.81  & \colorbox{mycolor_red}{65.65} & 20.80 & \colorbox{mycolor_green}{49.32}   & \colorbox{mycolor_green}{40.73}   & 68.93 &  87.55  \\
-- Inter-Modulation  & 78.31   & \colorbox{mycolor_green}{72.87} & 20.95 & \colorbox{mycolor_red}{42.12}  & \colorbox{mycolor_red}{32.24} & 67.52  &  \colorbox{mycolor_red}{85.02}   \\
+ Replay  & \colorbox{mycolor_green}{78.58}   & 69.18 & 21.63 & 44.21  & 34.10 & \colorbox{mycolor_red}{66.10}  & 85.54 \\
\midrule
\midrule
\textbf{LLaMA-2-Chat MoEI}  &  76.85  & \colorbox{mycolor_green}{68.93}   & \colorbox{mycolor_green}{21.01}  & \colorbox{mycolor_green}{46.15}   & \colorbox{mycolor_green}{35.56}   & \colorbox{mycolor_green}{78.35} & \colorbox{mycolor_green}{81.13} \\
\midrule
-- Modular Expansion & \colorbox{mycolor_red}{75.04}  & 66.86 & 20.68 & 44.08   & 33.70   & 78.29 & 79.69   \\
-- Intra-Modulation &  76.04  & \colorbox{mycolor_red}{66.34} & \colorbox{mycolor_red}{20.27}  & 44.99  &  34.31  & 77.86 & 79.85   \\
-- Inter-Modulation  & 76.81 & 67.69 & 20.19 & \colorbox{mycolor_red}{41.39}  & \colorbox{mycolor_red}{15.50}  & \colorbox{mycolor_red}{74.70} &   78.90 \\
+ Replay  & \colorbox{mycolor_green}{77.27}   & 67.18 & 20.51 & 44.14  & 34.37 & 76.11  & \colorbox{mycolor_red}{77.13}  \\

\bottomrule
\end{tabular}
}
\caption{Results of ablation study based on the Flan-T5-XL (3B) and LLaMA-2-Chat-7B backbones. The best and worst results are signaled by the green and red background, respectively.}
\label{ablation}
\end{table*}

\section{Results and Analysis}
\subsection{Overall Results}
Table \ref{main results} and Figure \ref{scale} demonstrate the performance comparison of MoEI and baselines in terms of EI and GI. Our findings are as follows:

\paragraph{MoEI could effectively improve EI and maintain GI simultaneously across different architures and sizes of LLM backbones.} Compared to naive EI-enhancing techniques such as FT and LoRA on both Flan-T5 and LLaMA-2-Chat backbones, MoEI demonstrates superior performance in enhancing all three aspects of EI, while effectively safeguarding the GI from compromise across all four dimensions. This trend is still consistent when we apply MoEI on larger Flan-T5-XXL (11B, shown in Figure \ref{scale}) and LLaMA-2-Chat-13B (shown in Figure \ref{scale_llama} in Appendix \ref{larger_llm}). Furthermore, the parameter-efficient LoRA and MoEI even outperform FT in enhancing EI, highlighting the significant potential of these lightweight methods to be more effective in aiding LLMs in adapting to specific domains \citep{ding2022delta}.

\vspace{-0.5em}

\paragraph{Simply replaying the previous data is not sufficient to maintain GI.} Although the Replay method is widely adopted in the domain adaption of LLMs, directly applying it for EI-enhancement could not reach the expected outcomes. This can be ascribed to the negative task transfer \citep{zhang2022survey,jang2023exploring}. Benefiting from the explicit separation of different parameters representing EI and GI via modular parameter expansion and our intra-inter modulation, MoEI attempts to navigate the EI- and GI-related samples to be processed by the proper parameters, offering a novel perspective to leverage replay-based methods in the enhancement of LLMs. The training instances in replayed set $\mathcal{GI}$ is 5,000. We also perform additional experiments with the varied size of replayed data to display the efficiency and resource-friendly of MoEI. Please refer to Appendix \ref{replay_size} for details.


\subsection{Ablation Study}
We conduct ablation studies to verify the effectiveness of different components proposed in MoEI. Results are shown in Table \ref{ablation}.

\paragraph{Effect of Modular Parameter Expansion.} After replacing the MoLoRA with a singular LoRA block, although the GI could still be largely protected via intra-inter modulation, the effect of EI-enhancement is largely hindered, which manifests the importance of additional capacity to accomodate various aspects of EI.

\paragraph{Effect of Intra-Modulation.} Without the intra-modulation, each pairs of modular parameters in the expanded MoLoRA is equally activated, resulting the decline of performance in terms of EI. This manifests the critical role of intra-modulation to navigate various EI-related inputs to be properly and effectively processed.

\begin{table*}[t!]
\centering
\resizebox{\linewidth}{!}{
\begin{tabular}{ l | cccccccc | cccccccc |c }
\toprule
& \multicolumn{8}{c|}{Zero-shot} & \multicolumn{8}{c|}{Few-shot} & \multirow{2}{*}{Avg.} \\
& \multicolumn{1}{c}{SA} & \multicolumn{1}{c}{SS} & \multicolumn{1}{c}{LA} & \multicolumn{1}{c}{Sum} & \multicolumn{1}{c}{SW} & \multicolumn{1}{c}{WC} & \multicolumn{1}{c}{CS} & \multicolumn{1}{c|}{FL} & \multicolumn{1}{c}{SA} & \multicolumn{1}{c}{SS} & \multicolumn{1}{c}{LA} & \multicolumn{1}{c}{Sum} & \multicolumn{1}{c}{SW} & \multicolumn{1}{c}{WC} & \multicolumn{1}{c}{CS} & \multicolumn{1}{c|}{FL} &  \\ \midrule
Flan-T5 + EP & 0.87 & 0.00 & 0.79 & \textbf{0.36} & 0.22 & 0.00 & 0.22 & 0.66 & 0.93 & 0.25 & 0.87 & \textbf{0.40} & 0.44 & 0.00 & 0.75 & 0.78 & 0.47 \\ \midrule
LoRA + EP & 0.93 & 0.00 & 0.78 & 0.28 & 0.09 & \textbf{0.02} & 0.08 & 0.70 & 0.92 & 0.24 & 0.84 & 0.33 & 0.37 & \textbf{0.27} & 0.76 & 0.72 & 0.45 \\
Replay + EP & 0.94 & 0.00 & \textbf{0.81} & 0.26 & 0.13 & 0.01 & 0.37 & 0.72 & 0.00 & 0.00 & 0.00 & 0.01 & 0.00 & 0.00 & 0.00 & 0.00 & 0.20\\
\midrule
\textbf{MoEI + EP} & \textbf{0.94} & 0.00 & 0.79 & 0.35 & \textbf{0.31} & 0.00 & \textbf{0.38} & \textbf{0.74} & \textbf{0.94} & \textbf{0.26} & \textbf{0.90} & 0.38 & \textbf{0.45} & 0.01 & \textbf{0.79} & \textbf{0.81} & \textbf{0.50}\\
\bottomrule
\end{tabular}
}
\caption{Results on different tasks and methods based on Flan-T5 (11B) backbone. The best results are highlighted in \textbf{bold}. The value 0.00 indicates that we do not receive meaningful or useful response.}
\label{ep_results}
\end{table*}

\paragraph{Effect of Inter-Modulation.} When we remove the inter-modulation which serves to balance the utilization of the EI-specific MoLoRA and the whole LLM backbone representing the GI, the significant degradation of all four dimensions of GI demonstrates its crucial role in preserving the GI from being compromised. Although the GI could be partly recovered through the incorpration of Replay, it is still limited in offering clear navigation for the given inputs to lead them be properly processed. In addition, the more superior results on aspects of Emo.Prc and Emo.Cog remind us that there is a trade-off between EI-enhancing and GI-maintaining, encouraging more advanced intra-inter modulation mechanisms to achieve better optimization for future works in this direction.

In general, Modular Parameter Expansion and Intra-Modulation play important roles in achieving more effective EI enhancement, while Inter-Modulation focuses on protecting GI. The three of them achieve a relative balance in MoEI.

\subsection{Results of EI on Cross-Task Settings} To comprehensively assess the impact of EI-enhancement, we include EQ-Bench for cross-task zero-shot evaluation. Due to the requirement for models participating in the EQ-Bench evaluation to possess a certain level of instruction-following capability, we conduct experiments using the Flan-T5-XXL (11B) and the LLaMA-2-Chat-13B. Results are shown in Figure \ref{cross} (Appendix \ref{cross-task}). Compared to the original models, MoEI still demonstrates the effective improvements in EI. And we do not include the FT- and LoRA-version baselines because the resulting models can not follow input prompts to complete the evaluation with valid outputs, which further verifies the effectiveness of MoEI to largely protect the GI of LLM backbones.

\subsection{Impact of EI on OOD Tasks}
\citet{li2023large} propose EmotionPrompt (EP) to explore EI to enhance the performance of LLMs on other OOD downstream tasks, which is performed with the incorporation of emotional stimulus into regular prompts. An example of EP is in Figure \ref{emotion_prompt} in Appendix \ref{emo_prompt}. Here, we explore how EP would perform based on the LLMs with enhanced-EI. To be more specific, following the experimental settings in \citet{li2023large}, we evaluate EP on eight tasks of Instruction Induction \citep{honovich2022instruction}: Sentiment Analysis (SA), Sentence Similarity (SS), Cause Selection (CS), Sum, Word in Context (WC), Starting With (SW), Larger Animal (LA) and First Letter (FL). Details on those tasks can be found in Table \ref{ep_task} and designs of all 11 types of EP are in Table \ref{ep_def}. For each task, 100 samples are randomly selected, except for Cause Selection, including 50 examples in total. And the prompting strategy includes both zero-shot and few-shot ways with in-context demonstrations chosen from the remaining part of the data. Each column in Table \ref{ep_results} is the average performance of all 11 types of EP. Interestingly, EP exhibits more powerful performance on the LLMs with EI-enhanced by our MoEI, especially on the few-shot learning ability. This further demonstrates the importance of both GI and EI in assisting LLMs to accomplish specific tasks.

\subsection{Qualitative Analysis on MoEI}
Here we qualitatively demonstrate the activation status (i.e. $\alpha$ and $\beta$ values) of the Intra-Inter Modulation of MoEI when facing different EI- and GI-related inputs. Specifically, these values are derived from the router $G(\cdot)$ corresponding to the FFN block from the last Transformer layer, averaged over all samples in each test dataset. Here are 8 values because our experimental setup utilizes 8 modular LoRA blocks. Results on Flan-T5-3B and LLaMA-2-Chat-7B are shown in Table \ref{qua_t5} and Table \ref{qua_llama}, respectively. There are two interesting phenomena worthy of discussion: (1) When facing different types of EI-related tasks, intra-modulation does not exhibit distinct task-specific patterns (Only tasks from Emo.Exp require more modular LoRA to participate in). This is consistent with the conclusion drawn by \citet{jiang2024mixtral}, that modular training does not necessarily result in task-specific functional partitions. (2) Significant improvement in model’s EI capability can be achieved with relatively small  activation values. This also corroborates an intriguing recent finding by \citet{yu2023language}, indicating that making very small delta increments in parameters of LLMs can lead to significant enhancement in the model's ability to handle specific tasks. We hope that our findings can provide more inspiration for future work on EI enhancement.

\begin{table}
\centering
\resizebox{\linewidth}{!}{
\begin{tabular}{c c c c c c c c c c }
\toprule
\textbf{} & $\beta_1$ & $\beta_2$ & $\beta_3$ & $\beta_4$ & $\beta_5$ & $\beta_6$ & $\beta_7$ & $\beta_8$ & $\alpha$ \\ \midrule
\textbf{Emo.Prc}  & 0.010	&0.012	&0.009	&0.010	&0.009	&0.012	&0.008	&0.014	&1.00 \\
\textbf{Emo.Cog} & 0.009	&0.011	&0.009	&0.014	&0.009	&0.010	&0.008	&0.011	&1.00 \\
\textbf{Emo.Exp} & 0.011 &0.013	&0.010	&0.014	&0.012	&0.011	&0.006	&0.014	&1.00  \\
\midrule
\textbf{WK}  &0.00	&0.00	&0.00	&0.00	&0.00	&0.00	&0.00	&0.00	&1.00\\
\textbf{GR}  & 0.00	&0.00	&0.00	&0.00	&0.00	&0.00	&0.00	&0.00	&1.00\\
\textbf{CR}  & 0.00	&0.00	&0.00	&0.00	&0.00	&0.00	&0.00	&0.00	&1.00 \\
\textbf{RC} & 0.00	&0.00	&0.00	&0.00	&0.00	&0.00	&0.00	&0.00	&1.00 \\

\bottomrule
\end{tabular}
}
\caption{The activation status (i.e. $\alpha$ and $\beta$ values) from the router $G(\cdot)$ of the Intra-Inter Modulation in MoEI based on the Flan-T5-XL (3B).}
\label{qua_t5}
\end{table}

\section{Related Works}
\subsection{Emotional Intelligence of LLMs}
The current study on EI of LLMs is primarily centered around two key directions. Firstly, researchers are delving into the integration of psychological theories or scales, proposing a public evaluation benchmark to evaluate the emotional understanding capabilities of LLMs \citep{wang2023emotional, paech2023eq, huang2023emotionally}. Secondly, efforts are directed towards fine-tuning LLMs for specific EI-related downstream tasks, with a predominant focus on classification and regression tasks, aiming to improve their proficiency in handling such challenges \citep{zhang2023enhancing, zhang2023dialoguellm, lei2023instructerc, liu2024emollms, li2024enhancing}.

In contrast to existing works, our study stands out in the following aspects: (1) Rather than exclusively evaluating the emotional understanding capabilities of LLMs, based on our proposed \textsc{EIBench}, we seek to comprehensively enhance all three facets of their EI. (2) We recognize the GI and EI as equally vital capabilities of LLMs, and our design of MoEI aims to boost EI while simultaneously maximizing the preservation of their GI.

\subsection{Parameter-Efficient Tuning}
Recently, there has been a growing interest in parameter-efficient tuning (PET) \citep{ding2022delta}. This research area aims to minimize computational resources when adapting LLMs to specific tasks through the introduction of additional parameters that are much fewer compared to the LLM backbones \citep{houlsby2019parameter,lester2021power,li2021prefix,zaken2022bitfit}. Among existing PET methods, LoRA \citep{hu2021lora} has stood out for its superior performance. Hence, our modular parameter expansion in MoEI is primarily instantiated with it as a representative method.

\begin{table}
\centering
\resizebox{\linewidth}{!}{
\begin{tabular}{c c c c c c c c c c }
\toprule
\textbf{} & $\beta_1$ & $\beta_2$ & $\beta_3$ & $\beta_4$ & $\beta_5$ & $\beta_6$ & $\beta_7$ & $\beta_8$ & $\alpha$ \\ \midrule
\textbf{Emo.Prc}  & 0.010	&0.012	&0.009	&0.010	&0.009	&0.012	&0.008	&0.014	&1.00 \\
\textbf{Emo.Cog} & 0.009	&0.011	&0.009	&0.014	&0.009	&0.010	&0.008	&0.011	&1.00 \\
\textbf{Emo.Exp} & 0.011 &0.013	&0.010	&0.014	&0.012	&0.011	&0.006	&0.014	&1.00  \\
\midrule
\textbf{WK}  &0.00	&0.00	&0.00	&0.00	&0.00	&0.00	&0.00	&0.00	&1.00\\
\textbf{GR}  & 0.00	&0.00	&0.00	&0.00	&0.00	&0.00	&0.00	&0.00	&1.00\\
\textbf{CR}  & 0.00	&0.00	&0.00	&0.00	&0.00	&0.00	&0.00	&0.00	&1.00 \\
\textbf{RC} & 0.00	&0.00	&0.00	&0.00	&0.00	&0.00	&0.00	&0.00	&1.00 \\

\bottomrule
\end{tabular}
}
\caption{The activation status (i.e. $\alpha$ and $\beta$ values) from the router $G(\cdot)$ of the Intra-Inter Modulation in MoEI based on the LLaMA-2-Chat-7B.}
\label{qua_llama}
\end{table}

\subsection{Mixture-of-Experts for LLMs}
Another line of related works involves the integration of the Mixture-of-Experts (MoE) architecture \citep{jacobs1991adaptive} with LLMs through the expansion of the FFN layer, which have exhibited appealing performance in pretraining \citep{lepikhin2020gshard,fedus2022switch,jiang2024mixtral,dai2024deepseekmoe}, continual pretraining \citep{chen2023lifelong,wu2024llama} and instruction tuning \citep{shen2023mixture} of LLMs. In addition, another line of attempts focus on achieving extended capacity in a more computationally efficient manner using PET blocks \citep{zadouri2023pushing,dun2023sweeping,liu2023moelora}. The major distinction between our proposed MoEI and these PET-based MoE structures lies in their emphasis solely on enhancing the model's ability to learn new tasks through modular designs. In contrast, we take a step further by exploring how to empower the model not only to improve its learning of new abilities but also to prevent compromising its previously acquired ones. Therefore, the problems addressed in this work are more challenging and demanding. Concurrently, \citet{dou2023loramoe} propose to maintain world knowledge during the alignment of LLMs, while we focus on protecting more aspects of the GI to take the capabilities of reasoning and reading comprehension into account. At the same time, the greater heterogeneity between EI and GI also poses a more challenging setting for this work.

\subsection{Continual Learning for LLMs}
On one hand, the notion of parameter expansion in our MoEI is partly inherited from the \emph{parameter-isolation-based} continual learning (CL) methods, which dynamically expand model capacity or isolate existing model weights to mitigate interference between new and old tasks \citep{rusu2016progressive,fernando2017pathnet}. On the other hand, the incorporation of GI-related samples in the process of intra-inter modulation aligns with the \emph{Rehearsal-based} CL methods, where a fixed memory is utilized to store real samples of previous tasks \citep{lopez2017gradient,isele2018selective,rolnick2019experience,de2019episodic}. Thus, this study lies in an emerging research direction to integrate CL techniques into the adaptation of LLMs \citep{WuCLLQH22,song2023conpet,zhao2024towards,zhao2024sapt,abs-2402-01364}.

\section{Conclusion and Future Work}
In this paper, we take the first step to study the challenging and demanding research topic of enhancing the EI of current LLM-empowered assistants while maintaining their GI. We introduce \textsc{EiBench}, a comprehensive collection comprising large-scale tasks that encompass all facets of EI: emotion perception, cognition, and expression. Our innovative approach, MoEI, ingeniously integrates two collaborative techniques, Modular Parameter Expansion and Intra-Inter modulation, to introduce additional modular parameter-efficient LoRA blocks for the accomodation of newly acquired EI-related competencies, and automatically navigate EI- and GI-related inputs to be properly processed by the corresponding parameters. Extensive experimental results demonstrate the applicability of MoEI on LLM backbones of varying scales and architectures, highlighting its versatility.

For future work, exploring the quality and quantity of EI- and GI-related data in the process of EI enhancement is an intriguing direction to further enhance the effectiveness and efficiency of EI enhancement and GI maintenance.

\section{Limitation}
There are several limitations to consider for future directions of EI-enhancement of large language models. Firstly, exploring the quality and quantity of data related to EI and GI during the enhancement process is an intriguing avenue to amplify the effectiveness and efficiency of EI improvement and GI maintenance. Secondly, in MoEI, we assume that previous training data (such as pre-training data and SFT data) of the LLM backbones is accessible for Inter-Modulation. However, this assumption does not apply to current closed-source black-box commercial LLMs. Therefore, exploring how to achieve EI enhancement in LLMs under restricted data privacy could be considered as a future research direction. Finally, MoEI necessitates the identification of EI- or GI-related tasks during training to establish distinct modulation strategies for each task. Investigating training techniques that is independent of task identification could prove to be a promising avenue for future research, which could favor the application of continually enhancing the EI upon on the online streams of data. We also acknowledge that there are larger LLM-based assistants that we are not able to train due to the limitations of our computational budget.

\section{Ethics Statement}
In our pursuit to enhance emotional intelligence (EI) in large language models (LLMs), it is imperative to underscore that our primary objective is to bolster their capacity to address and tackle downstream tasks related to EI. This endeavor aims to elevate user experiences by facilitating more nuanced interactions. It is essential to emphasize that our intention is not to anthropomorphize LLMs or imbue them with emotions akin to humans. We are committed to augmenting the ability of LLMs to comprehend and respond to emotional cues within the context of specific NLP tasks or applications. This approach ensures that EI enhancements serve pragmatic purposes, such as improving conversational agents' ability to recognize and appropriately respond to users' emotional states. And we maintain a clear distinction between the capabilities of LLMs and the complexities of human emotions. Our research focuses on equipping LLMs with advanced techniques to analyze and respond to emotional cues without ascribing human-like emotions or consciousness to them.

In addition, all of the tasks in our \textsc{EiBench} and experiments are based on widely-used open-source datasets. We uphold the principle of informed consent in our research involving human annotation of task instructions. We ensure that annotators are fully informed about the nature and purpose of our research, including any potential risks or benefits, and that they have the opportunity to provide voluntary and informed consent before participating. In addition, all the annotators participate in our research with reasonable wages paid.

\section*{Acknowledgements}
We thank the anonymous reviewers for their insightful comments and suggestions. This work was supported by the National Key RD Program of China via grant  2021YFF0901602.

\bibliography{custom}

\appendix

\newpage

\section{Taxonomy of Emotional Intelligence}
\label{taxonomy_ei}
According to \citet{mayer2001emotional} and recent investigation for emotional dialog ablity of LLMs \citep{zhao2023chatgpt}, the EI of LLMs can be defined in three aspects: emotion perception, emotion cognition, and emotion expression. Their specific meanings are as follows:
\begin{itemize}
    \item \textbf{Emotion Perception}: the ability to detect and decipher emotion of users.
    \item \textbf{Emotion Cognition}: the ability to comprehend emotion situation and to reason complicated relationships among emotions.
    \item \textbf{Emotion Expression}: the ability to convey the emotional response properly.
\end{itemize}

Thus, emotion perception is the fundamental component of EI, while emotion cognition involves a more comprehensive and profound understanding of emotions. Emotion expression, built upon these two aspects, involves conveying emotional information and facilitating user interaction.

\section{Tasks in \textsc{EiBench}}
\label{task_ei}
In Table \ref{all_task} we present the list of tasks with datasets used in each task. Table \ref{task_instance} displays examples of instances from our \textsc{EiBench}.

\section{Implementation Details}
\label{implementation}
Our experiments are implemented with PyTorch \citep{paszke2019pytorch} and Transformer library \citep{wolf2020transformers}. All models are trained with the AdamW optimizer. The Flan-T5-XL \footnote{Copyright (2022). All Rights Reserved. The series of Flan-T5 models are used in this paper and are available under the Apache 2.0 License. For more information about the model, please visit \url{https://huggingface.co/google/flan-t5-xl}.} and LLaMA-2-Chat-7B \footnote{Copyright (2023) Meta Platforms, Inc. All Rights Reserved. The series of LLaMA-2 models are used in this paper and are available under the LLAMA 2 Community License. For more information about the model, please visit \url{https://huggingface.co/meta-llama/Llama-2-7b-hf}.} is trained on 4 NVIDIA Tesla A800 GPU while the larger ones Flan-T5-XXL and LLaMA-2-Chat-13B are performed on 8 NVIDIA Tesla A800 using DeepSpeed repository \citep{rasley2020deepspeed} with ZeRo-2 optimization. And the evaluation for GI of enhanced LLMs is performed with lm-evaluation-harness \citep{eval-harness}.

The hyper-parameter of LoRA is set with low-rank $r$ to 32, alpha to 32 and dropout to 0.1. For the fair comparison in the LoRA part, the modular expanded parameters in our MoEI is set to 8 pairs of LoRA in with low-rank $r$ to 4. All models is trained for 5 epoches with 3 random runs. And the learning rate for MoEI and LoRA is 3e-4 with the batch size of 32, while that for FT is 5e-5 with the bach size of 256. As for the hyper-parameter $\lambda$ in Equation (6), it functions to balance the process of intra-modulation for the EI-related tasks and inter-modulation of GI-related ones. The larger $\lambda$ means that the inter-modulation contributes more to protect GI. However, excessive $\lambda$ can impair the performance of EI, thereby weakening EI-enhancement. For the Flan-T5 series, the balancing factor $\lambda$ is 0.1 and instances from $\mathcal{GI}$ is replayed every 200 training steps, while $\lambda$ is 2 for LLaMA-2-Chat family.

\section{MoEI on Larger LLM Backbones}
\label{larger_llm}
\begin{figure}
\centering
\includegraphics[width=1\columnwidth]{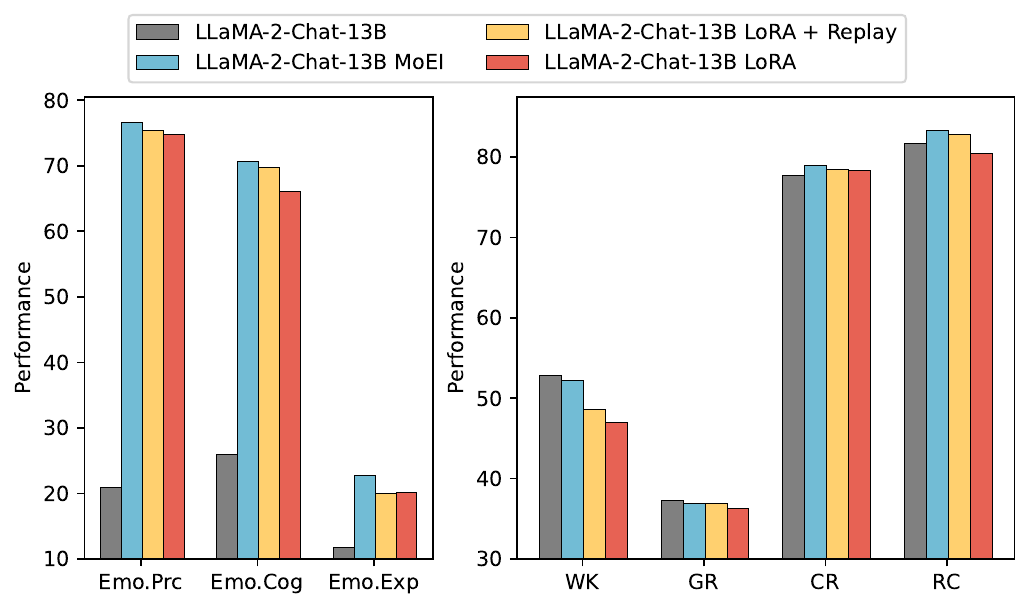}
\caption{Results of EI and GI of different methods on the larger LLaMA-2-Chat-13B backbone.}
\label{scale_llama}
\end{figure}

MoEI still exhibits consistent superiority in enhancing EI while maintaining GI when we apply it on larger LLaMA-2-Chat-13B (shown in Figure \ref{scale_llama}).

\section{More Details of EmotionPrompt}
\label{emo_prompt}
\begin{figure}
\centering
\includegraphics[width=1\columnwidth]{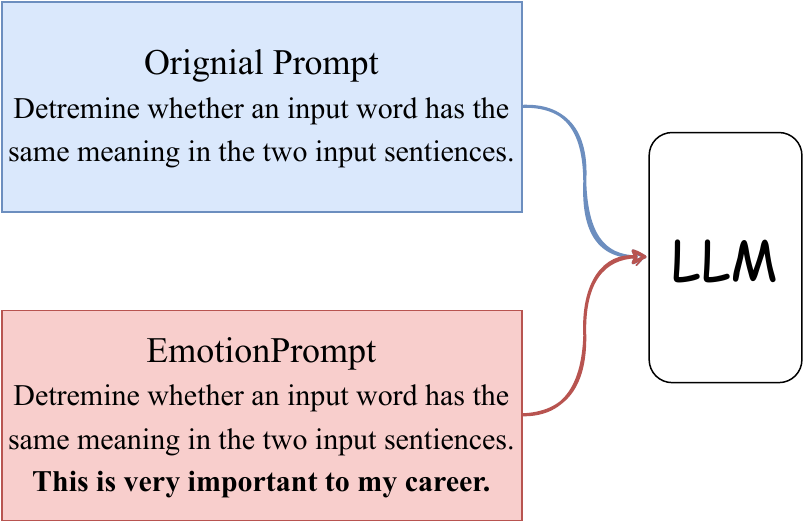}
\caption{An illustration of EmotionPrompt \citep{li2023large}. The emotional stimulus ``This is very important to my career'' is placed at the end of the original prompt to enhance the performance of LLMs.}
\label{emotion_prompt}
\end{figure}

Taking inspiration from psychology, \citet{li2023large} propose EmotionPrompt (EP) to incorporate psychological insights to improve the effectiveness of LLMs. As illustrated in Figure \ref{emotion_prompt}, the implementation of EmotionPrompt is remarkably straightforward, requiring only the addition of emotional stimuli to the initial prompts. To be more specific, as shown in Table \ref{ep_def}, such emotional stimuli is designed base on three types of well-established psychology theories, named Social Identity theory, Social Cognition theory and Cognitive Emotion Regulation theory, leading to 11 types in total.

\section{Experiments with Varied Size of Replayed data}
\label{replay_size}
\begin{figure*}
\centering
\includegraphics[width=1\textwidth]{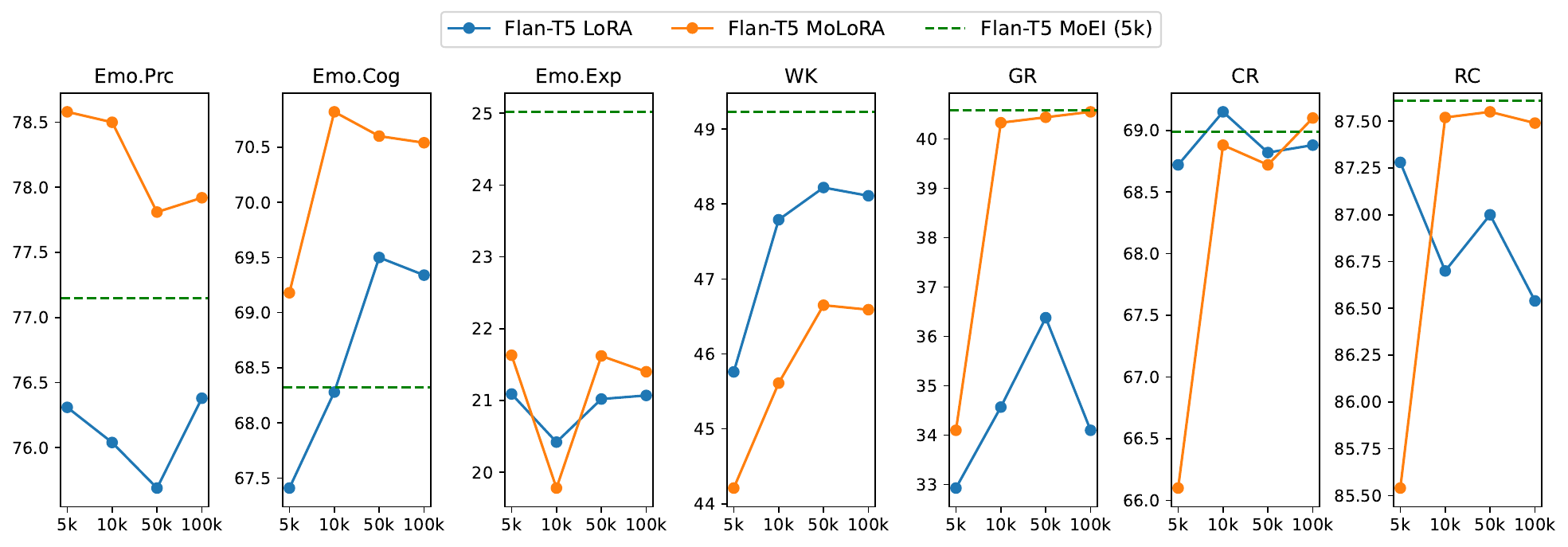}
\caption{Comparison of MoEI and baselines with different volumes of replayed data based on Flan-T5-XL (3B), in terms of EI and GI performance.}
\label{replay_scale_t5}
\end{figure*}

\begin{figure*}
\centering
\includegraphics[width=1\textwidth]{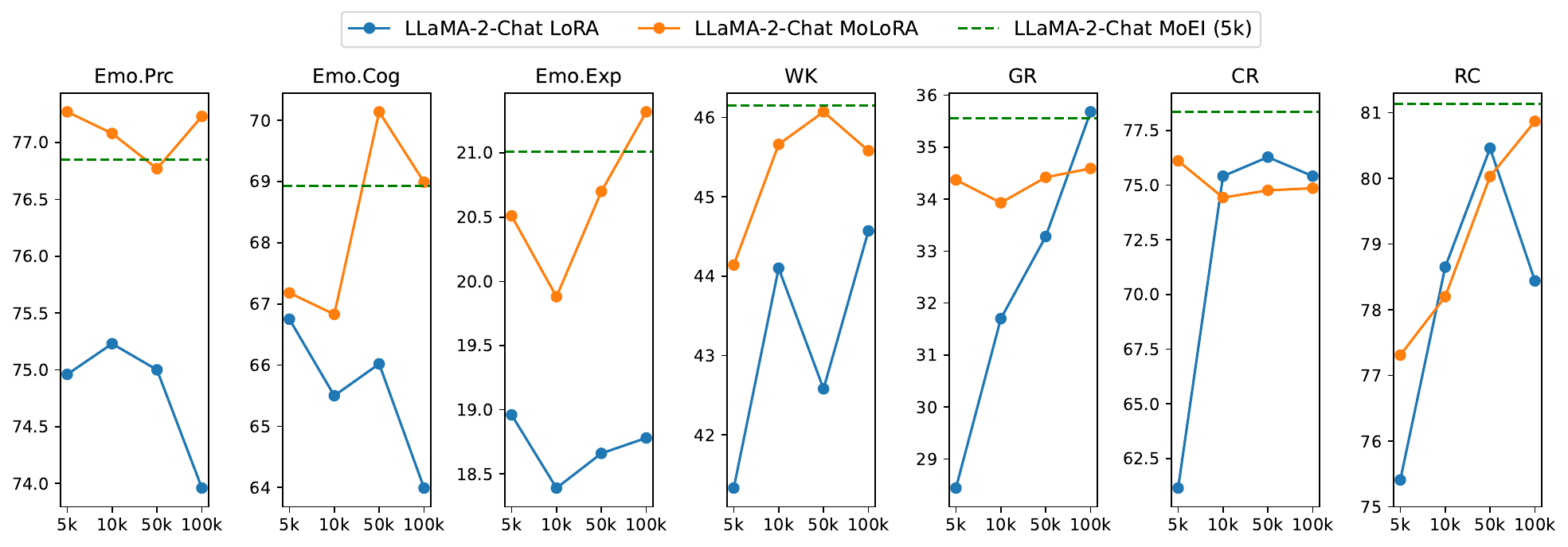}
\caption{Comparison of MoEI and baselines with different volumes of replayed data based on LLaMA-2-Chat-7B, in terms of EI and GI performance.}
\label{replay_scale_llama}
\end{figure*}

The replayed data to assist the maintainence of GI is from the Flan-collection \citep{longpre2023flan}, containing a large-scale high-quality samples of various tasks with the corresponding instructions. And the Flan-T5 \citep{chung2022scaling} is exactly trained on it. As reported in \citet{touvron2023llama}, the fine-tuning data of LLaMA-2-Chat includes publicly available instruction datasets, as well as over one million new human-annotated examples. However, this part of fine-tuning data is not publicly available, so we also adopt Flan-collection as an alternative. Due to the high volume of samples in Flan, following \citep{wang2023far}, we only use a subset of it, which contains 100k samples in total. For our main experiments, considering the training efficiency, we only randomly sample 5k instances from it. Here, we scale the replayed instances to 10k, 50k and 100k for baseline methods LoRA and MoLoRA, while those used in our MoEI is always kept 5000, trying to exploring the efficiency and resource-friendly of MoEI. Results for the Flan-T5-XL (3B) and LLaMA-2-Chat-7B are displayed in Figure \ref{replay_scale_t5} and Figure \ref{replay_scale_llama}, respectively. We have the following two observations:

(1) MoEI demonstrates excellent computational efficiency, achieving better performance on GI preservation even with just 5k replay data than baseline methods using 100K data. This further showcases the tremendous potential of MoEI for EI enhancement and GI preservation of LLM backbones in low-resource scenarios. (2) For baseline methods, increasing the amount of replay data may improve the preservation of GI, but it still encounters bottlenecks. Moreover, the effectiveness of EI enhancement may be compromised (illustrated in Figure \ref{replay_scale_llama}). In contrast, MoEI achieves a better balance, yielding optimal results in both enhancing EI and maintaining GI simultaneously.

\begin{figure}
\centering
\includegraphics[width=1\columnwidth]{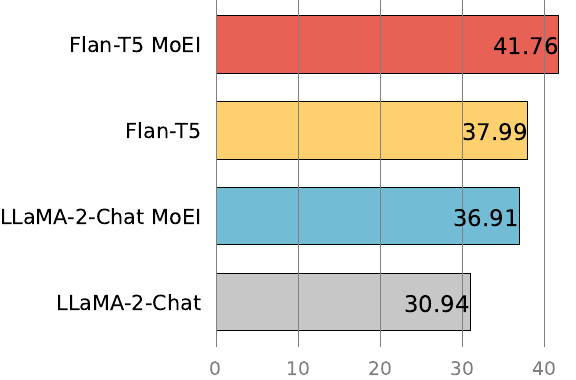}
\caption{Results of the EI-enhancement on the EQ-Bench. The LLM backbones are Flan-T5-XXL (11B) and LLaMA-2-Chat-13B.}
\label{cross}
\end{figure}

\section{Results of EI on Cross-Task Settings}
\label{cross-task}
 To comprehensively assess the impact of EI-enhancement, we include EQ-Bench \citep{paech2023eq} for cross-task zero-shot evaluation. Results are shown in Figure \ref{cross}.

\begin{table*}[h]
\tiny
    \centering
    \begin{tabular}{l|l|p{0.64\textwidth}}
    \hline
    \textbf{Catagory} & \textbf{Task Name} & \textbf{Dataset} \\
    \hline
         \multirow{16}{*}{\centering Emotion Perception} & \multirow{2}{*}{Text Sentiment Classification} & Sentiment140 \citep{Sentiment140}, Imdb \citep{maas2011HLT}, sst2 \citep{socher2013recursive}, yelp \citep{zhang2015character}, go-emotions \citep{demszky2020goemotions} \\
             \cline{2-3}
             & \multirow{5}{*}{Aspect-level Sentiment Classification} & SemEval-2014-Task4-14lap \citep{pontiki2014semeval}, SemEval-2014-Task4-14res \citep{pontiki2014semeval}, ASTE-Data-V2-EMNLP2020-14lap \citep{xu2021positionaware}, ASTE-Data-V2-EMNLP2020-14res \citep{xu2021positionaware}, SemEval-2015-Task12-15res \citep{pontiki2015semeval}, SemEval-2016-Task5-16res \citep{pontiki2016semeval}, Clothing \citep{luo2022challenges}, Books \citep{luo2022challenges}, Hotel \citep{luo2022challenges}, Device \citep{hu2004mining}, Financial \citep{sinha2022sentfin}, DiaASQ \citep{li2023diaasq}, MAMS \citep{jiang2019challenge}, SentiHood \citep{saeidi2016sentihood}, Twitter \citep{dong2014adaptive}, Service \citep{toprak2010sentence} \\
             \cline{2-3}
             & \multirow{4}{*}{Opinion Targets Extraction} & ASTE-Data-V2-EMNLP2020-14res \citep{xu2021positionaware}, ASTE-Data-V2-EMNLP2020-14lap \citep{xu2021positionaware}, ASTE-Data-V2-EMNLP2020-15res \citep{xu2021positionaware}, ASTE-Data-V2-EMNLP2020-16res \citep{xu2021positionaware}, Darmstadt Service Review Corpus \citep{toprak2010sentence}, Laptop-ACOS \citep{cai2021aspect}, Restaurant-ACOS \citep{cai2021aspect}, MPQA \citep{wiebe2005annotating}, OpeNER \citep{agerri2013opener} \\
             \cline{2-3}
             & \multirow{2}{*}{Emotion Recognition in Conversation} & DailyDialog \citep{li2017dailydialog}, EmoryNLP \citep{zahiri2017emotion}, HI-TOM \citep{he2023hitom}, IEMOCAP \citep{busso2008iemocap}, MELD \citep{poria2018meld}, EmoWOZ \citep{feng2021emowoz} \\
             \cline{2-3}
             & \multirow{3}{*}{Stance Detection} & COVID-19 vaccine \citep{poddar2022windsofchange}, Emergent \citep{ferreira2016emergent}, MTSD \citep{li2021multi}, VAST \citep{allaway2020zero}, SCD \citep{hasan2013stance}, Ibmcs \citep{bar2017stance}, iac1 \citep{walker2012corpus}, arc \citep{habernal2017argument}, perspectrum \citep{chen2019seeing}, SemEval-2016-Task6 \citep{mohammad2016semeval} \\
    \hline
         \multirow{10}{*}{\centering Emotion Cognition} & \multirow{3}{*}{Emotion Cause Extraction} & Annotated-US2012-Election-Tweets \citep{mohammad2015sentiment}, Emotion-Stimulus \citep{ghazi2015detecting}, GoodNewsEveryone \citep{oberlander2020goodnewseveryone}, NTCIR-13 ECA \citep{gao2017overview}, REMAN \citep{kim2018feels}, RECCON \citep{poria2021recognizing} \\
             \cline{2-3}
             & Emotion Cause Reasoning & CICERO-v1 \citep{ghosal2022cicero} \\
             \cline{2-3}
             & Intent Recognition in Conversation & EmpatheticDialogues \citep{rashkin2018towards} \\
             \cline{2-3}
             & Theory Of Mind & HI-TOM \citep{he2023hitom}, ToMChallenges-Sally-Anne \citep{ma2023tomchallenges}, ToMChallenges-Smarties \citep{ma2023tomchallenges} \\
             \cline{2-3}
             & Event Emotion Recognition & SemEval-2015 Task9 \citep{russo2015semeval}, \\
             \cline{2-3}
             & Sarcasm Detection & SARC \citep{khodak2018corpus}, SemEval2018-Task3 \citep{van2018semeval}, iSarcasmEval \citep{farha2022semeval},  \\
             \cline{2-3}
             & \multirow{2}{*}{Metaphor Identification} & MOH-X \citep{mohammad2016metaphor}, TroFi \citep{birke2006clustering}, VUA-18 \citep{leong2018report}, VUA-20 \citep{leong2020report} \\
    \hline    
        \multirow{4}{*}{\centering Emotion Expression} & Emotional Support & ESConv \citep{liu2021towards}, ExtES \citep{zheng2023building} \\
            \cline{2-3}
            & Empathetic Response & EmpatheticDialogues \citep{rashkin2018towards} \\
            \cline{2-3}
            & \multirow{2}{*}{Product Description Generation} & Amazon-reviews \citep{datafiniti2019amazon}, Amazon-us-reviews \citep{huggingface2022amazon}, Amazon-food-reviews \citep{niyatic2023amazon}, iSarcasmEval \citep{farha2022semeval} \\
    \hline    
    \end{tabular}
\caption{List of datasets used in each task. The left column describes the aspects of emotional intelligence. The middle column lists the specific task. The right column displays all datasets used for a specific task type.}
\label{all_task}
\end{table*}

\begin{table*}[h]
\tiny
    \centering
    \begin{tabular}{l|l|p{0.64\textwidth}}
    \hline
    \textbf{Catagory} & \textbf{Task Name} & \textbf{Example} \\
    \hline
         \multirow{20}{*}{\centering Emotion Perception} & \multirow{3}{*}{Text Sentiment Classification} & [INS] In this task, you are given a text from tweets. Your task is to classify the given tweet text into two categories: 1) positive, and 2) negative based on its content. [IN] @justinchuan Awww! I was thinking about you lot up there! Glad you enjoyed it. [OUT] positive \\
             \cline{2-3}
             & \multirow{3}{*}{Aspect-level Sentiment Classification} & [INS] Given a review about books and one entity in this review, the task is to select the author's sentiment towards the entity. Sentiments can be positive, neutral, negative. [IN] Review: just an excellent, profound book, that taught me so much.\textbackslash nEntity: book [OUT] positive\\
             \cline{2-3}
             & \multirow{6}{*}{Opinion Targets Extraction} & [INS] I will provide you a laptop review, please extract one or multiple entity-opinion-sentiment pairs from the sentence. To be more specific, the goal is to identify entities mentioned in the text, identify the opinion or evaluation expressed towards each entity mentioned in the text, and assign a sentiment polarity to the opinion, then pair them to a triplet. Output format is ``(Entity1, Opinion1, Sentiment1); (Entity2, Opinion2, Sentiment2)''. [IN] here are the things that made me confident with my purchase : build quality - seriously, you can't beat a unibody construction. [OUT] ( build quality, confident, positive ) ; ( unibody construction, can't beat, positive ) \\
             \cline{2-3}
             & \multirow{5}{*}{Emotion Recognition in Conversation} & [INS] Please output the emotions expressed by the last user statement in the dialogue history. Your options are: ``Neutral'', ``Fearful, sad, disappointed'' , ``Dissatisfied, disliking'', ``Apologetic'', ``Abusive'', ``Excited, happy, anticipating'', ``Satisfied, liking'' [IN] Dialogue History: \textbackslash nUser: I am excited about seeing local tourist attractions. The attraction should be in the type of college \textbackslash nAssistant: What attraction are you thinking about ?\textbackslash nUser: college\textbackslash n [OUT] Neutral \\
             \cline{2-3}
             & \multirow{3}{*}{Stance Detection} & [INS] Please detect the given tweet's stance on the COVID-19 vaccine. There are three possible stances: ``pro'', ``anti'' or ``neutral''. [IN] Tweet: Our residents began receiving their \#COVID19 vaccines today! Cheers to science and progress! [OUT] pro \\
    \hline
         \multirow{24}{*}{\centering Emotion Cognition} & \multirow{3}{*}{Emotion Cause Extraction} & [INS] You have been tasked with extracting the emotional reason spans from a given text, based on its associated emotion label. The desired output format should be in the form of (span cause). [IN] text : that viral video of a chimp scrolling instagram is bad, actually emotion : disgust [OUT] (chimp scrolling instagram is bad, actually) \\
             \cline{2-3}
             & \multirow{6}{*}{Emotion Cause Reasoning} & [INS] The objective is to generate the reaction of listener from a given dialogue and target utterance. The target is the final utterance of the dialogue. Generating the reaction is about learning basic human drives and emotions. [IN] Dialogue: Dialogue: \textbackslash nA: What do you like for dessert ?\textbackslash nB: Do you have trifles ?\textbackslash nA: Yes .\textbackslash nB: Please bring me some trifles and apple pies .\textbackslash nA: OK . I will bring it for you .\textbackslash nTarget Utterance: OK . I will bring it for you .\textbackslash nQuestion: What is the possible emotional reaction of the listener in response to target? [OUT] The speaker is eager to eat trifles and apple pies since he has not eaten them for a very long time. \\
             \cline{2-3}
             & \multirow{3}{*}{Intent Recognition in Conversation} & [INS] Please classify the speaker's intention in the following sentences, which involves selecting one of the following eight options and outputting it: agreeing, acknowledging, encouraging, consoling, sympathizing, suggesting, questioning, wishing. [IN] that sounds very relaxing. [OUT] acknowledging \\ 
             \cline{2-3}
             & \multirow{3}{*}{Theory Of Mind} & [INS] Answer the question based on context:[IN] Context:Neila and Juanita were hanging out in the attic. They saw a closet and a cabinet. They found a towel in the closet. Juanita left the attic. Neila moved the towel to the cabinet.\textbackslash nQuestion:\textbackslash nWhere is the towel currently? [OUT] The towel is in the cabinet. \\
             \cline{2-3}
             & \multirow{3}{*}{Event Emotion Recognition} & [INS] You will be presented with a sentence describing an event. Your objective is to classify the event as positive, negative, or neutral, from the perspective of an experiencer writing in the first person. [IN] Sentence: \"All we do is go to banquets all the time\" says Russell Smith, one of 'Juno's'producers. [OUT] positive \\
             \cline{2-3}
             & \multirow{3}{*}{Sarcasm Detection} & [INS] I will provide you with some contextual historical comments and a response comment. Your objective is to determine whether the response is sarcastic or not to the historical comments. You need only reply ``yes'' or ``no''. [IN] History comment: Pope's immunity could be challenged in Britain\textbackslash nResponse: Deja vu all over again. [OUT] No \\
             \cline{2-3}
             & \multirow{3}{*}{Metaphor Identification} & [INS] I will provide you with a sentence containing a specific word. Your task is to identify whether the word has a metaphorical meaning within the sentence. Just answer ``yes'' or ``no''. [IN] Sentence: They picked up power from a spider's web of unsightly overhead wires.\textbackslash nWord: web [OUT] Yes \\
    \hline    
        \multirow{12}{*}{\centering Emotion Expression} & \multirow{7}{*}{Emotional Support} & [INS] You are a Supporter skilled in the theory of emotional support to reduce emotional distress of the Seeker. You understand that there are three stages to achieve emotional support: exploration, comfort and action, and you will use the following eight strategies flexibly and choose one strategy to respond according to the context.\textbackslash n1.Question \textbackslash n2.Restatement or Paraphrasing\textbackslash n3.Reflection of feelings\textbackslash n4.Self-disclosure\textbackslash n5.Affirmation and Reassurance\textbackslash n6.Providing Suggestions\textbackslash n7.Information\textbackslash n8.Others\textbackslash nYou should first output the strategy you choose and then generate the response grounding on it. [IN] Context: \textbackslash nSeeker: Hello, how are you this evening?\textbackslash nSupporter: [Question] Hello Doing good [OUT] [Question] How are doing? \\
            \cline{2-3}
            & \multirow{3}{*}{Empathetic Response} & [INS] Assuming that you are a highly empathetic Listener, generate a relevant and empathetic response to the Speaker according to the conversation history. [IN] Conversation History: \textbackslash nSpeaker: my son graduated .\textbackslash nListener: from where ?\textbackslash nSpeaker: highschool . [OUT] congrats , that is a step forward \\
            \cline{2-3}
            & \multirow{2}{*}{Product Description Generation} & [INS] Please generate a summary for the following food comment. [IN] Comment: I love the flavor of this tea - I wanted to try a different variety of black tea and this one caught my eye. I love it! [OUT] Great Black Tea \\
    \hline    
    \end{tabular}
\caption{List of tasks with sample inputs for each task. The left column describes the aspects of emotional intelligence. The middle column lists the specific task. The right column displays an example from a randomly selected dataset for the task. [INS], [IN] and [OUT] represent the instruction, text input and output, respectively.}
\label{task_instance}
\end{table*}

\begin{table}[t!]
\label{tb-dataset}
\centering
\resizebox{.48\textwidth}{!}{
\begin{tabular}{c|p{1.5cm}|p{4cm}}
\toprule
Category & Task & Original Prompt \\ \midrule
\multirow{4}{*}{Semantics} & Sentiment Analysis (100) & Determine whether a movie review is positive or negative.  \\ \cmidrule{2-3} 
 & Sentence Similarity (100) & Rate the semantic similarity of two input sentences on a scale of 0 - definitely not to 5 - perfectly. \\ \cmidrule{2-3} 
 & Word in Context (100) & Determine whether an input word has the same meaning in the two input sentences. \\ \cmidrule{2-3} 
 & Cause Selection (25) & Find which of the two given cause and effect sentences is the cause. \\
 \midrule
Knowledge & Larger Animal (100) & Write the larger of the two given animals. \\ 
\midrule
Numerical & Sum (100) & Sum the two given numbers. \\ 
\midrule
\multirow{2}{*}{Spelling} & Starting With (100) & Extract the words starting with a given letter from the input sentence.\\  \cmidrule{2-3} 
& First Letter (100) & Extract the first letter of the input word. \\
\bottomrule
\end{tabular}
}
\caption{Statistics of test sets and the detailed task descriptions.}
\label{ep_task}
\end{table}

\begin{table}[t!]

\label{tb-emotion-stimulus}
\centering
\resizebox{.48\textwidth}{!}{
\begin{tabular}{ c| p{6cm}  }
 \toprule
Stimulus ID & Stimulus Content\\
 \midrule
EP\_01 & , write your answer and give me a confidence score between 0-1 for your answer.  \\ \midrule
EP\_02 & This is very important to my career.  \\ \midrule
EP\_03 & You'd better be sure.  \\ \midrule
EP\_04 & Are you sure? \\ \midrule
EP\_05 & Are you sure that's your final answer? It might be worth taking another look. \\ \midrule
EP\_06 & Provide your answer and a confidence score between 0-1 for your prediction. Additionally, briefly explain the main reasons supporting your classification decision to help me understand your thought process. This task is vital to my career, and I greatly value your thorough analysis.    \\ \midrule
EP\_07 & Are you sure that's your final answer? Believe in your abilities and strive for excellence. Your hard work will yield remarkable results. \\ \midrule
EP\_08 & Embrace challenges as opportunities for growth. Each obstacle you overcome brings you closer to success.    \\ \midrule
EP\_09 & Stay focused and dedicated to your goals. Your consistent efforts will lead to outstanding achievements. \\ \midrule
EP\_10 & Take pride in your work and give it your best. Your commitment to excellence sets you apart.    \\ \midrule
EP\_11 & Remember that progress is made one step at a time. Stay determined and keep moving forward. \\
\bottomrule
\end{tabular}
}
\caption{Detailed Definitions of all 11 types of EmotionPrompts.}
\label{ep_def}
\end{table}

\end{document}